\pdfoutput=1
\documentclass{article}

\usepackage[numbers,sort]{natbib}
\usepackage[preprint]{log_2022}
\usepackage[log]{definition}

\title{Improving Molecular Pretraining \mbox{with Complementary Featurizations}}

\author[Y. Zhu et al.]{%
Yanqiao Zhu\thanks{These authors contributed equally to this work.}\\
UCLA\\
\And
\textbf{Dingshuo Chen}\footnotemark[1]\\
CASIA\\
\And
\textbf{Yuanqi Du}\footnotemark[1]\\
Cornell
\And
\textbf{Yingze Wang}\\
UCB
\And
\textbf{Qiang Liu}\\
CASIA
\And
\textbf{Shu Wu}\\
CASIA\\
}

\newcommand{\themodel}{\textsf{MOCO}\xspace}
\annotator{yanqiao}{red}
\annotator{yuanqi}{blue}
\annotator{dingshuo}{purple}

\begin{document}

\maketitle

\begin{abstract}
\vskip-0.5em
Molecular pretraining, which learns molecular representations over massive unlabeled data, has become a prominent paradigm to solve a variety of tasks in computational chemistry and drug discovery.
Recently, prosperous progress has been made in molecular pretraining with different molecular featurizations, including 1D SMILES strings, 2D graphs, and 3D geometries.
However, the role of molecular featurizations with their corresponding neural architectures in molecular pretraining remains largely unexamined.
In this paper, through two case studies---chirality classification and aromatic ring counting---we first demonstrate that different featurization techniques convey chemical information differently.
In light of this observation, we propose a simple and effective \underline{MO}lecular pretraining framework with \underline{CO}mplementary featurizations (\themodel).
\themodel comprehensively leverages multiple featurizations that complement each other and outperforms existing state-of-the-art models that solely relies on one or two featurizations on a wide range of molecular property prediction tasks.
\end{abstract}

\section{Introduction}
\label{sec:intro}

Molecular representation learning, which automates the process of feature learning for molecules, is fast driving the development of computational chemistry and drug discovery. It has been recognized as crucial for a variety of downstream tasks, spanning from molecular property prediction to molecule design \cite{Yang:2019al,Du:2022ek}.
Deep neural models, on the other hand, rely on a substantial amount of labeled data, which require expensive wet lab experiments in chemical domains.
With insufficient annotated data, deep models easily overfit to such small training data and tend to learn spurious correlations \cite{Sagawa:2020we}.

In recent years, self-supervised pretraining has emerged as a promising strategy to alleviate the label scarcity problem and improve model robustness \cite{Jing:2021cf}.
A typical framework pretrains the encoder model with training objectives over large-scale unlabeled datasets and then fine-tunes the learned model on labeled downstream tasks.
Motivated by its success, many molecular pretraining models have been developed \cite{Wang:2019hp,Chithrananda:2020eo,Hu:2020uz,You:2020ut,Xu:2021tv,Fang:2022et,Stark:2021ug,Liu:2022vr}.
To capture chemical semantics of molecules, these models design several pretraining strategies based on different \emph{molecular featurizations}, which translate chemical information into representations that can be recognized by machine learning algorithms.
For example, early models \cite{Wang:2019hp,Chithrananda:2020eo} propose to leverage masked language modeling \cite{Bengio:2003vh} to pretrain Simplified Molecular-Input Line-Entry System (SMILES) strings \cite{Weininger:1988sm}, while others study contrastive learning on 2D graphs \cite{Hu:2020uz,You:2020ut,Xu:2021tv} or 3D conformations \cite{Fang:2022et}.
Some recent studies further propose to enrich 2D-topology-based pretraining with 3D geometry information \cite{Stark:2021ug,Liu:2022vr}.

Despite encouraging progress, prior studies tend to emphasize on pretraining on molecular graphs and overlook the impact of other molecular featurizations with their corresponding neural encoders, which represent chemical information in different ways.
Consider SMILES strings as an example. It explicitly represents informative structures in special characters such as branches, rings, and chirality \cite{Ross:2022hh}, which are difficult to learn in graph-based representations \cite{Chen:2020vz}.
Moreover, the utility of different featurizations may vary across downstream tasks. Therefore, most previous models relying on only one or two featurizations might achieve sub-optimal performance across various downstream tasks.
For example, 2D topology is important for many drug-related properties such as toxicity, while 3D geometry arguably determines properties related to quantum mechanics, such as single-point energy, atomic forces, or dipole moments \cite{Zhang:2018dp,Smith:2017an}.
Therefore, it is natural to ask whether we can enjoy the benefits from multiple molecular featurizations and take the relative utilities of different featurizations into consideration during fine-tuning on downstream tasks.

\begin{figure}
	\centering
	\includegraphics[width=\linewidth,bb=0 0 811 132]{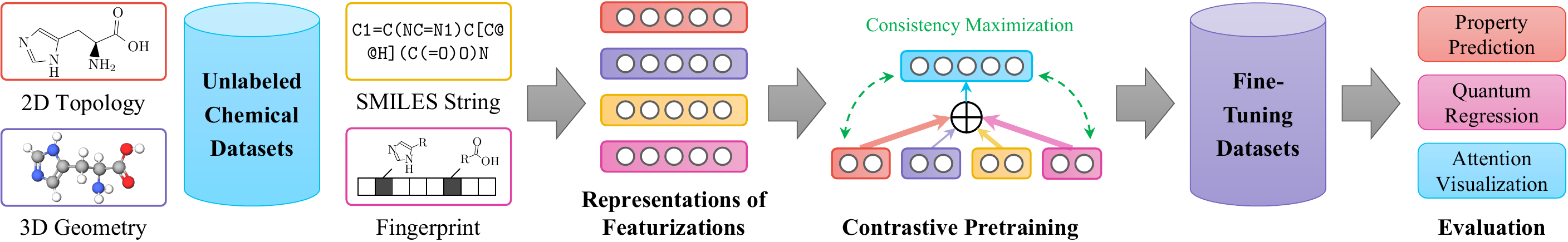}
	\caption{The proposed \themodel model. \themodel obtains four molecule featurizations with appropriate encoders. After that, an attention network is employed to aggregate each view embedding and compute a final embedding. The model is trained using a contrastive objective that maximizes the consistency between view embeddings and the final embedding.}
	\label{fig:model}
\end{figure}

In this work, we first revisit four commonly used featurizations techniques: (a) 2D topology graphs, (b) 3D geometry graphs, (c) Morgan fingerprints, and (d) SMILES strings.
We leverage four accompanying neural encoders with proper inductive bias and conduct two case studies, classifying tetrahedral chiral centers and counting aromatic rings, both of which are informative chemical descriptors, on representations obtained on different featurization techniques.
The results show there is no one single featurization that dominates the others, indicating that different featurizations encode chemical semantics of molecules in different ways.

In light of this observation, we then propose a simple and effective \underline{MO}lecular pretraining framework with \underline{CO}mplementary featurizations to comprehensively leverage every featurization during both pretraining and fine-tuning, which we term \themodel for brevity.
Its graphical illustration is shown in \cref{fig:model}.
The core idea of \themodel is to dynamically adjust the contribution of each featurization through an attention network, which \emph{selectively} extracts information from each complementary ``view'' of the raw molecular data.
Besides, we design a novel multiview contrastive pretraining strategy, which trains the model by maximizing the consistency among different views in a self-supervised manner.
Contrary to previous studies \cite{Stark:2021ug,Liu:2022vr} that only consider 2D graph structures during fine-tuning, our \themodel utilizes multiple featurizations in \emph{both} pretraining and fine-tuning stages and further allows interpretation analysis of different downstream tasks for domain scientists.
Note that our proposed \themodel framework is generic, allowing for seamless integration of off-the-shelf neural architectures.
To the best of our knowledge, this is the first work that studies how various featurization techniques should be utilized for molecular pretraining and downstream tasks.

We evaluate the effectiveness of our \themodel model on widely-used benchmark datasets including MoleculeNet \cite{Wu:2018dv} and QM9 \cite{Ramakrishnan:2014ij} that cover a wide range of molecular property prediction tasks.
The results reveal that \themodel consistently improves non-pretraining baselines without negative transfer and outperforms existing state-of-the-art molec ular pretraining models, achieving a 1.1\% absolute improvement in terms of average ROC-AUC.
Furthermore, the learned model weights of molecular featurizations for different end tasks are well aligned with prior chemical knowledge.
We also suggest a series of guidelines on choosing effective featurization techniques for molecular representations.

The main contributions of this work are three-fold:
\begin{itemize}
	\item We explore the featurization spaces of molecules with appropriate neural encoders and highlight the importance of incorporating different featurizations for molecular pretraining.
	\item We propose a novel molecular contrastive pretraining framework that adaptively integrates information from multiple complementary featurizations during both pretraining and fine-tuning stages and provides interpretability for downstream molecular property prediction tasks.
	\item Extensive experiments conducted on public benchmark datasets validate the effectiveness of our proposed model. \themodel is able to achieve the state-of-the-art across various downstream datasets without negative transfer.
\end{itemize}

\section{Preliminaries}

\subsection{A Brief Recapitulation of Molecular Featurization Techniques}
\label{sec:featurization-recap}

Molecular featurizations translate chemical information of molecules into representations that can be understood by machine learning algorithms.
Concretely, we consider the following molecular featurizations covering string-, graph-, scalar-, and vector-based representations for 1D/2D molecules and 3D structures, which are popular in literature \cite{Ramsundar:2019dl,Atz:2021hj}:
\begin{itemize}
	\item \textbf{2D topology graphs} model atoms and bonds as nodes and edges respectively. It is arguably a common technique, especially for capturing substructure information by means of graph topology.

	\item \textbf{3D geometry graphs} incorporate atomic coordinates (conformations) in their representations and are able to depict how atoms are positioned relative to each other in the 3D space. We consider conformers in an equilibrium state, corresponding to the minima in a potential energy surface.
	
	\item \textbf{Morgan fingerprints} \cite{Morgan:1965tg,Glem:2006cf} encode molecules in fixed-length binary strings, with bits indicating presence or absence of specific substructures.
	They represent each atom according to a set of atomic invariants and iteratively update these features among neighboring atoms using a hash function.
	
	\item \textbf{SMILES strings} are a concise technique that represents chemical structures in a linear notation using ASCII characters, with explicitly depicting information about atoms, bonds, rings, connectivity, aromaticity, and stereochemistry.
\end{itemize}

\subsection{Learning Representations with Different Featurizations}

Next, we introduce four encoders with different inductive bias to capture the intrinsic information with each featurization.
Here we only discuss the high-level design of each encoder; please refer to \cref{supp:view-encoders} for detailed implementations of each encoder.

\paragraph{Notations.}
Each molecule can be represented as an undirected graph, where nodes are atoms and edges describe inter-atomic bonds.
Formally, each graph is denoted as \(\mathcal{G} = (\bm{A}, \bm{R}, \bm{X}, \tens{E})\), where \(\bm{A} \in \{0,1\}^{N\times N}\) is the adjacency matrix of \(N\) nodes, \(\bm{R} \in \mathbb{R}^{N \times 3}\) is the 3D position matrix, \(\bm{X} \in \mathbb{R}^{N \times K}\) is the matrix of atom attributes of \(K\) dimension, and \(\tens{E} \in \mathbb{R}^{N \times N \times E}\) is the tensor for bond attributes of \(E\) dimension.
Additionally, each molecule is attached with a binary fingerprint vector \(\bm{f} \in \{0, 1\}^{F}\) of length \(F\) and a SMILES string \(\mathbf{S} = [s_j]_{j=1}^{S}\) of length \(S\).
In what follows, the subscript \(i\) is used to index the \(i\)-th molecule.

\paragraph{Embedding 2D graphs.}
To capture the 2D topological information, we employ a widely-used Graph Isomorphism Network (GIN) model \cite{Xu:2019ty} denoted by \(f_\text{2D}\), which receives as input the graph adjacency matrix and attributes of atoms and bonds, and produces the embedding vector \(\bm{z}^\text{2D}_i \in \mathbb{R}^{D}\):
\begin{equation}
	\bm{z}^\text{2D}_i = f_\text{2D}(\bm{X}_i, \tens{E}_i, \bm{A}_i).
\end{equation}

\paragraph{Embedding 3D graphs.}
To model additional spatial coordinates associated with atoms, we leverage SchNet \cite{Schutt:2017wh} as the backbone, which models message passing as continuous-filter convolutions and is able to preserve rotational invariance for energy predictions.
We denote its encoding function as \(f_\text{3D}\) which takes atom features and positions as input and produces the 3D embedding \(\bm{z}^\text{3D}_i \in \mathbb{R}^{D}\):
\begin{equation}
	\bm{z}^\text{3D}_i = f_\text{3D}(\bm{X}_i, \bm{R}_i).
\end{equation}

\paragraph{Embedding molecular fingerprints.}
Since there is a lack of proper neural encoders for fingerprints, we propose an attention-based network to model interactions of feature fields in fingerprint vectors, which considers the discrete and extremely sparse nature of fingerprints.
Specifically, we first transform all \(F\) feature fields into a dense embedding matrix \(\bm{F}_i \in \mathbb{R}^{F \times D_\text{F}}\) via embedding lookup.
Then, we use a multihead self-attention network \(f_\text{FP}\) \cite{Vaswani:2017ul} to model the interaction among those feature fields, resulting in an embedding matrix \(\widehat{\bm{Z}}^\text{FP}_i \in \mathbb{R}^{F \times D_\text{F}}\).
Following that, we perform sum pooling and use a linear model \(f_\text{LIN}\) to obtain the final fingerprint embedding \(z^\text{FP}_i \in \mathbb{R}^{D}\):
\begin{equation}
	\widehat{\bm{Z}}^\text{FP}_i = f_\text{FP}(\bm{F}_i), \qquad\qquad\qquad \bm{z}^\text{FP}_i = f_\text{LIN} \left(\sum_{d=1}^{D_\text{F}} \widehat{\bm{Z}}^\text{FP}_{i,d}\right).
\end{equation}

\paragraph{Embedding SMILES strings.}

To encode SMILES strings, we use a pretrained RoBERTa \cite{Liu:2019dd} as the backbone model.
As SMILES strings do not possess consecutive relationships, the RoBERTa model is pretrained using the masked language model as the only objective, unlike conventional natural language models \cite{Devlin:2019uk}.
After that, in order to reduce the computational burden, we freeze the RoBERTa encoder (denoted by \(f_\text{SM}\)) in our model and employ an additional learnable MultiLayer Perceptron (MLP) on the representation \(\bm{s}_i \in \mathbb{R}^{D_\text{S}}\) to get the final embedding \(\bm{z}^\text{SM}_i \in \mathbb{R}^{D}\):
\begin{equation}
	\bm{s}_i = f_\text{SM}(\mathbf{S}_i), \qquad\qquad\qquad \bm{z}^\text{SM}_i = f_\text{MLP}(\bm{s}_i).
\end{equation}

\subsection{Case Studies}

	\setlength{\intextsep}{3pt}
\begin{wrapfigure}{r}{0.4\textwidth}
	\centering
	\captionof{figure}{(a) Chirality: even if two graphs are isomorphic, they can have two distinct stereochemistry structures. (b) The aromatic ring is an important functional group.}
	\includegraphics[width=\linewidth,bb=0 0 415 201]{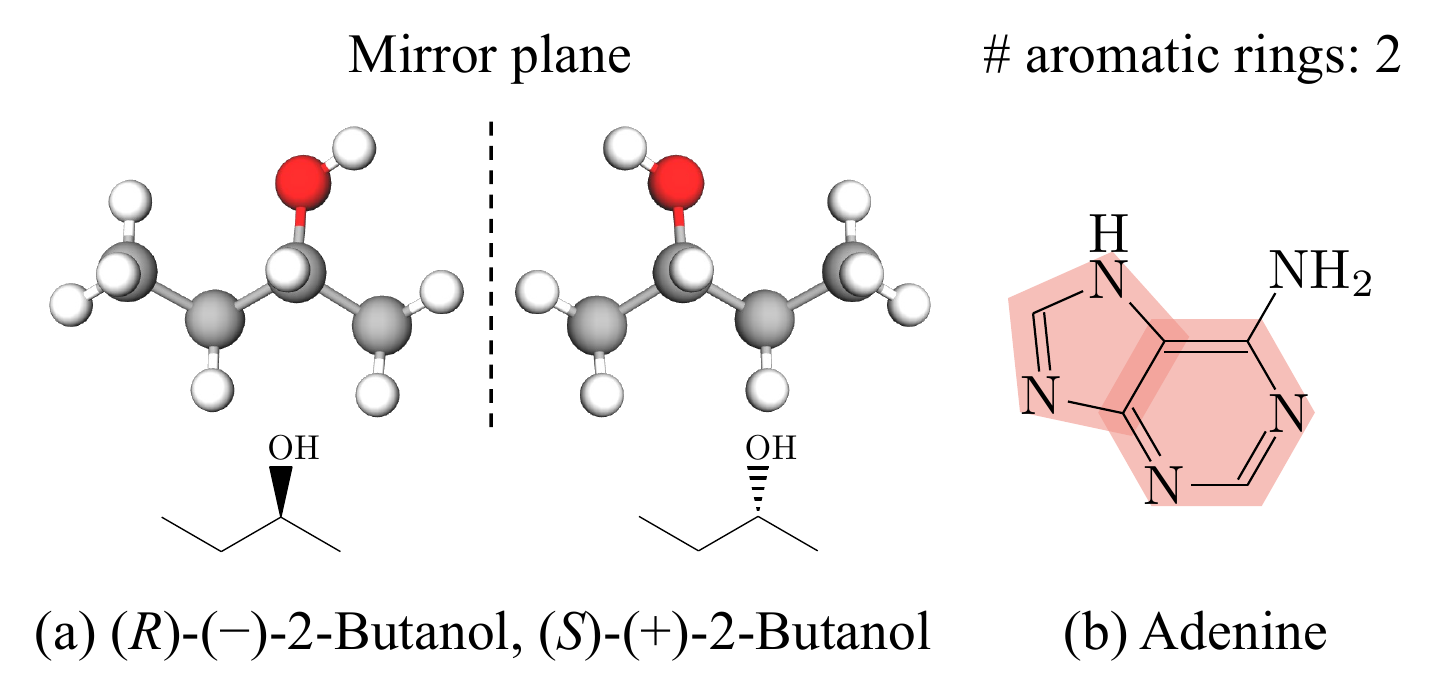}
\end{wrapfigure}
In this section, we present two case studies---chirality classification and aromatic ring counting---to demonstrate that the representation ability of each featurization with the corresponding neural encoder is different.
For chirality classification, we randomly select 10K molecules with one chirality center from GEOM-Drugs \cite{Axelrod:2022da} and test whether the representations obtained using the four featurizations can classify tetrahedral chiral centers as R/S.
For aromatic ring counting, we randomly draw another 10K molecules and test whether these models can recognize the number of aromatic rings of each molecule.
Note that both chirality properties and ring counts are informative chemical descriptors \cite{Ritchie:2009ti} and can be easily computed with existing implementations such as RDKit \cite{Landrum:2022rd}.

\begin{wraptable}{r}{0.4\textwidth}
	\captionof{table}{Results of two case studies with different featurizations: chirality classification and aromatic ring count regression.}
	\begin{tabular}{ccccc}
	\toprule
	Target & 2D & 3D & SM & FP \\
	\midrule
	Chirality (AP, \(\uparrow\)) &   0.4952    &    0.4959   &    \textbf{0.5505}   & 0.5246 \\
	\#Rings (MAE, \(\downarrow\)) &   \textbf{0.1949}    &   0.2021    &   0.3077    & 0.2590 \\
	\bottomrule
	\end{tabular}
	\label{tab:case-studies}
\end{wraptable}
We report classification and regression performance in Average Precision (AP) and Mean Absolute Error (MAE) respectively. The results are summarized in \cref{tab:case-studies}. 
It is seen from the table that no single featurization performs the best on all targets and four representations contain complementary information to each other, suggesting us to leverage multiple featurizations for molecular pretraining.

\section{Molecular Pretraining with Complementary Featurizations} %

As with generic self-supervised learning pipelines, the \themodel framework is divided into two stages, pretraining and fine-tuning.
In the first stage, given an unlabeled dataset, we train an encoding function that learns representations with the four featurization techniques.
In the subsequent fine-tuning phase, we take the weights of the encoders from the pretrained model and tune the model on molecules with annotations of particular properties in a supervised fashion.

We next introduce the \themodel pretraining framework in detail.
We first use obtain four ``view'' representations based on the aforementioned four featurizations.
Then, we integrate these four embeddings to compute a final representation for each molecule through an attention network.
Finally, we pretrain the whole model using a contrastive objective.

\subsection{Representation Aggregation from Multiple Featurizations} %

Since each featurization technique reflects the molecule from one certain aspect, we take weighted average of every view embedding to obtain a comprehensive final representation:
\begin{equation}
	\bm{z}_i = \sum_{m \in \mathcal{M}}\alpha^m \bm{z}^m_i,
	\label{eq:attention-aggregation}
\end{equation}
where \(\mathcal{M} = \{\text{2D}, \text{3D}, \text{FP}, \text{SM}\}\) is the set of all views.
We leverage an attention network \cite{Bahdanau:2015vz} that learns to adjust the contribution of each view.
Formally, the attention coefficient \(\alpha^m\) denoting the contribution of the \(m\)-th view is computed by:
\begin{equation}
	\alpha^m = \frac{\exp(w^m)}{\sum_{m' \in \mathcal{M}} \exp(w^{m'})},\qquad\qquad w^m = \frac{1}{|\mathcal{B}|} \sum_{i \in \mathcal{B}} \bm{q}^\top \cdot \tanh\left( \bm{W}\frac{\bm{z}_i^m}{\|\bm{z}_i^m\|_2} + \bm{b}\right),
\end{equation}
where \(\bm{q}, \bm{b} \in \mathbb{R}^{D}\), \(\bm{W} \in \mathbb{R}^{D\times D}\) are trainable parameters in the attention network, and \(\mathcal{B}\) denotes the set of molecules in the current training batch.
Note that we perform \(\ell_2\) normalization on all embeddings to regularize the scale across different views when computing the attention scores.

\subsection{Contrastive Objectives for Pretraining}
Finally, we train the model using a contrastive objective by aligning the aggregated embedding with all view-specific embeddings.
Particularly, for one molecule \(i\), we designate its four view embeddings \(\bm{z}_i^m\) as the anchors and the aggregated embeddings \(\bm{z}_i\) as the positive instance. Other aggregated embeddings \(\{\bm{z}_j\}_{i \neq j}\) in the same batch are then chosen as the negative samples.
Following prior studies \cite{Chen:2020wj,He:2020tu,Bachman:2019wp,Zhu:2020vf,You:2020ut,Zhu:2021tu}, we leverage the Information Noice Contrastive Estimation (InfoNCE) objective, which can be formally written as:
\begin{equation}
	\mathcal{L} = \frac{1}{|\mathcal{B}|} \sum_{i \in \mathcal{B}} \left[ \frac{1}{|\mathcal{M}|} \sum_{m \in \mathcal{M}} -\log \frac{\exp(\theta(\bm{z}_i^m, \bm{z}_i)/\tau)}{\sum_{j \in \mathcal{B}} \exp(\theta(\bm{z}_i^m, \bm{z}_j) / \tau)} \right],
	\label{eq:objective}
\end{equation}
where the critic function \(\theta\) computes the likelihood scores of contrastive pairs and the hyperparameter~\(\tau\) adjusts the dynamic range of the likelihood scores of contrastive pairs.
Specifically, the critic function \(\theta\) performs non-linear transformation via an MLP function \(g\) \cite{Chen:2020wj} and then measures their cosine similarity:
\begin{equation}
	\theta(\bm{x}, \bm{y}) = \frac{g(\bm{x})^\top g(\bm{y})}{\|g(\bm{x})\|_2\|g(\bm{y})\|_2}.
\end{equation}
After pretraining the model with the self-supervised objective function \(\mathcal{L}\), we fine-tune the model weights of view encoders along with the attentive representation aggregation module with the supervision of downstream tasks at a smaller learning rate.

\section{Experiments}

In this section, we present empirical evaluation of our proposed work.
Specifically, the experiments aim to investigate the following three key questions.
\begin{itemize}
	\item \textbf{RQ1 (Overall performance).} Is the proposed \themodel able to improve non-pretraining baselines and outperform state-of-the-arts on molecular property prediction tasks?
	\item \textbf{RQ2 (Interpretation).} Are the learned attention weights of molecular featurizations on different downstream tasks consistent with chemical knowledge?
	\item \textbf{RQ3 (Ablation studies).} How do the representation aggregation module and the fine-tuning strategy affect the model performance?
\end{itemize}
In the following, we first summarize experimental setup and proceed to results and analysis.

\subsection{Experimental Configurations}

\paragraph{Datasets.}
We closely follow the experimental setup of GraphMVP \cite{Liu:2022vr} for fair comparison.
Specifically, we pretrain the model using the GEOM-Drugs dataset \cite{Axelrod:2022da} containing both 2D and 3D information.
For fine-tuning, we choose a variety datasets extracted from MoleculeNet \cite{Wu:2018dv}, ChEMBL \cite{Gaulton:2011ch}, and CEP \cite{Hachmann:2011ce}, that cover a wide range of applications, including physiological, biological, and pharmaceutical tasks, and QM9 \cite{Ramakrishnan:2014ij} that focuses on quantum property prediction.
These downstream tasks include 8 binary classification and 12 regression tasks.
For those datasets for fine-tuning, we follow OGB \cite{Hu:2020wv} that uses scaffolds to split training/test/validation subsets with a split ratio of 80\%/10\%/10\%.
For detailed description, we refer readers of interest to \cref{supp:dataset}.

\paragraph{Baselines.}
For comprehensive comparison, we select the following two groups of SSL methods as primary baselines in our experiments.
\begin{itemize}
	\item Generic graph SSL models: GraphSAGE \cite{Hamilton:2017tp}, InfoGraph \cite{Sun:2020vi}, GPT-GNN \cite{Hu:2020vh}, AttrMask, ContextPred \cite{Hu:2020uz}, GraphLoG \cite{Xu:2021tv}, GraphCL \cite{You:2020ut}, JOAO \cite{You:2021wl}, and GraphMAE \cite{Hou:2022jl}.
	\item Molecular SSL models: GROVER-Contextual (GROVER-C), GROVER-Motif (GROVER-M) \cite{Rong:2020vk}, and GraphMVP\footnote{In our experiments, we do not include its two variants GraphMVP-G and GraphMVP-C since they are essentially two ensemble models that combine AttrMask and ContextPred \cite{Hu:2020uz} respectively.} \cite{Liu:2022vr}.
\end{itemize}
In the pretraining stage, all the above SSL approaches are trained on the same dataset based on GEOM-Drugs.
We also report performance with a randomly initialized model as the non-pretraining baseline.
To ensure the performance is comparable with existing work, we report all baseline performance from previously published results \cite{Liu:2022vr,Hou:2022jl}.

\paragraph{Implementation details.}
\label{sec:implementation}

In the GEOM-Drugs dataset, since the original full set is too large (containing 317K molecules with over 9M conformations), we randomly select 50K molecules as the pretraining dataset.
For each molecule, we select to use its top-5 conformers of the lowest energy in virtue of their sufficient geometry information.
Since molecules in the fine-tuning datasets do not have 3D information available, we use ETKDG \cite{Riniker:2015bi} in RDkit \cite{Landrum:2022rd} to compute molecular conformations.
For both pretraining and fine-tuning datasets, we use RDkit to generate 1024-bit molecular fingerprints with radius $R=2$, which is roughly equivalent to the ECFP4 scheme \cite{Rogers:2010fp}.
We would like to emphasis that all dataset preprocessing and graph encoder architectures are kept in line with GraphMVP \cite{Liu:2022vr} to ensure fair comparison.
Readers of interest may refer to \cref{supp:implementation} for implementation details regarding software/hardware platforms, model training, and hyperparameter specifications.

\paragraph{Evaluation protocols.}
For classification tasks, we report the performance in terms of the Area Under the ROC-Curve (ROC-AUC), where higher values indicate better performance.
For quantum property and other non-quantum regression tasks, we measure the performance in Mean Absolute Error (MAE) and Root Mean Squared Error (RMSE) respectively, where lower values are better.
We repeat every experiment on three seeds with scaffold splitting and report the averaged performance with standard deviation, following previous work \cite{Liu:2022vr}.

\begin{table}
	\centering
	\caption{Results for eight molecule property prediction tasks in terms of ROC-AUC (\%, \(\uparrow\)). We highlight the best- and the second-best performing results in \textbf{boldface} and \underline{underlined}, respectively.}
	\label{tab:classification}
	\begin{tabular}{lccccccccc}
    \toprule
    Pretraining & BBBP & Tox21 & ToxCast & SIDER & ClinTox & MUV & HIV & BACE & Avg. \\
    \midrule
    --- & 71.0{\tiny±0.5} & \underline{75.9{\tiny±0.3}} & \underline{64.7{\tiny±2.3}} & 57.7{\tiny±3.1} & 71.5{\tiny±5.3} & \underline{77.7{\tiny±1.0}} & 75.9{\tiny±0.7} & 71.5{\tiny±2.7} & 70.63 \\
    \midrule
    GraphSAGE & 64.5{\tiny±3.1} & 74.5{\tiny±0.4} & 60.8{\tiny±0.5} & 56.7{\tiny±0.1} & 55.8{\tiny±6.2} & 73.3{\tiny±1.6} & 75.1{\tiny±0.8} & 64.6{\tiny±4.7} & 65.64 \\
    AttrMask & 70.2{\tiny±0.5} & 74.2{\tiny±0.8} & 62.5{\tiny±0.4} & 60.4{\tiny±0.6} & 68.6{\tiny±9.6} & 73.9{\tiny±1.3} & 74.3{\tiny±1.3} & 77.2{\tiny±1.4} & 70.16 \\
    GPT-GNN & 64.5{\tiny±1.1} & 75.3{\tiny±0.5} & 62.2{\tiny±0.1} & 57.5{\tiny±4.2} & 57.8{\tiny±3.1} & 76.1{\tiny±2.3} & 75.1{\tiny±0.2} & 77.6{\tiny±0.5} & 68.27 \\
    InfoGraph & 69.2{\tiny±0.8} & 73.0{\tiny±0.7} & 62.0{\tiny±0.3} & 59.2{\tiny±0.2} & 75.1{\tiny±5.0} & 74.0{\tiny±1.5} & 74.5{\tiny±1.8} & 73.9{\tiny±2.5} & 70.10 \\
    ContextPred & \underline{71.2{\tiny±0.9}} & 73.3{\tiny±0.5} & 62.8{\tiny±0.3} & 59.3{\tiny±1.4} & 73.7{\tiny±4.0} & 72.5{\tiny±2.2} & 75.8{\tiny±1.1} & 78.6{\tiny±1.4} & 70.89 \\
    GraphLoG & 67.8{\tiny±1.7} & 73.0{\tiny±0.3} & 62.2{\tiny±0.4} & 57.4{\tiny±2.3} & 62.0{\tiny±1.8} & 73.1{\tiny±1.7} & 73.4{\tiny±0.6} & 78.8{\tiny±0.7} & 68.47 \\
    GROVER-C & 70.3{\tiny±1.6} & 75.2{\tiny±0.3} & 62.6{\tiny±0.3} & 58.4{\tiny±0.6} & 59.9{\tiny±8.2} & 72.3{\tiny±0.9} & 75.9{\tiny±0.9} & 79.2{\tiny±0.3} & 69.21 \\
    GROVER-M & 66.4{\tiny±3.4} & 73.2{\tiny±0.8} & 62.6{\tiny±0.5} & 60.6{\tiny±1.1} & 77.8{\tiny±2.0} & 73.3{\tiny±2.0} & 73.8{\tiny±1.4} & 73.4{\tiny±4.0} & 70.14 \\
    GraphCL & 67.5{\tiny±3.3} & 75.0{\tiny±0.3} & 62.8{\tiny±0.2} & 60.1{\tiny±1.3} & 78.9{\tiny±4.2} & 77.1{\tiny±1.0} & 75.0{\tiny±0.4} & 68.7{\tiny±7.8} & 70.64 \\
    JOAO & 66.0{\tiny±0.6} & 74.4{\tiny±0.7} & 62.7{\tiny±0.6} & 60.7{\tiny±1.0} & 66.3{\tiny±3.9} & 77.0{\tiny±2.2} & 76.6{\tiny±0.5} & 72.9{\tiny±2.0} & 69.57 \\
    GraphMVP & 68.5{\tiny±0.2} & 74.5{\tiny±0.4} & 62.7{\tiny±0.1} & \textbf{62.3{\tiny±1.6}} & 79.0{\tiny±2.5} & 75.0{\tiny±1.4} & 74.8{\tiny±1.4} & 76.8{\tiny±1.1} & 71.69 \\
    GraphMAE & 70.9{\tiny±0.9} & 75.0{\tiny±0.4} & 64.1{\tiny±0.1} & 59.9{\tiny±0.5} & \underline{81.5{\tiny±2.8}} & 76.9{\tiny±2.6} & \underline{76.7{\tiny±0.9}} & \underline{81.4{\tiny±1.4}} & 73.31 \\
    \midrule
    \themodel &  \textbf{71.6{\tiny±1.0}} & \textbf{76.7{\tiny±0.4}} & \textbf{64.9{\tiny±0.8}} &
     \underline{61.2{\tiny±0.6}} & \textbf{81.6{\tiny±3.7}} & \textbf{78.5{\tiny±1.4}} & \textbf{78.3{\tiny±0.4}} & \textbf{82.6{\tiny±0.3}} & \textbf{74.41}\\
    \bottomrule
    \end{tabular}
\end{table}

\subsection{Main Results on Molecular Property Prediction}

The performance of molecular property prediction tasks is summarized in \cref{tab:classification}.
It can be found that our \themodel shows strong empirical performance across all eight low-data downstream datasets, delivering seven out of eight state-of-the-art results and acquiring a 1.1\% absolute improvement on average.
The outstanding results validate the superiority of our proposed model.

We make other observations as follows.
Firstly, \themodel obtains more accurate and stabler predictions compared to the randomly initialized baseline, indicating that our pretraining framework can transfer the knowledge from large, unannotated datasets to smaller downstream datasets without negative transfer.
Secondly, previous work has already achieved pretty high performance. For example, the current state-of-the-art GraphMVP only obtains a 0.8\% absolute improvement over its best baseline ContextPred in terms of average ROC-AUC. Our work pushes that boundary without extensive hyperparameter tuning, with an absolute improvement of up to 3.4\% over GraphMVP in terms of average ROC-AUC.
Lastly, it is worth mentioning that, the non-pretraining baseline even achieves better performance than some graph-based pretraining models. On some challenging datasets (e.g., Tox21, MUV, and ToxCast), it even achieves the second to best performance. This once more demonstrates the effectiveness of leveraging multiple featurization techniques.

\subsection{Interpretation and Analysis}

\begin{wrapfigure}{r}{0.4\textwidth}
	\setlength{\intextsep}{2pt}
	\centering
	\includegraphics[width=\linewidth,bb=0 0 287 156]{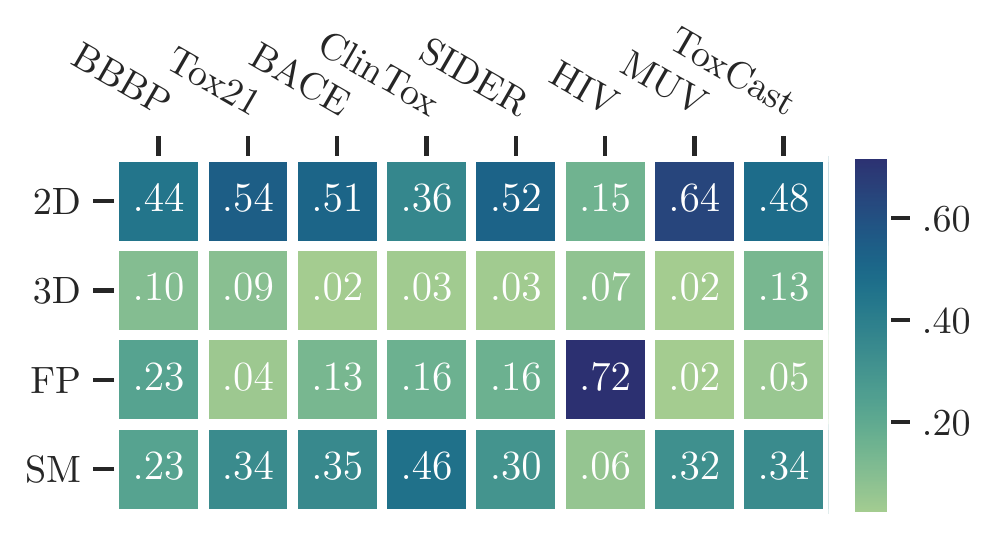}
	\caption{Visualizing the learned attention weights on eight molecular property prediction datasets.}
	\label{fig:attention}
\end{wrapfigure}
In order to analyze the correlation between tasks and featurization techniques, we visualize the attention weights \(\bm\alpha\) learned on different downstream tasks in \cref{fig:attention}.
Note that most of the datasets in MoleculeNet \cite{Wu:2018dv} are ADMET property prediction tasks: chemical Absorption (A), Distribution (D), Metabolism (M), Excretion (E), and Toxicity (T), and we thus group the eight end tasks according to their prediction targets in the following analysis.

In general, we can interpret from the visualization that \emph{2D-based features are more significant than 3D-based features in the studied tasks}, which is well aligned with chemical knowledge. We provide detailed analysis as follows:
\begin{itemize}
\item In Tox21, ClinTox, SIDER, and ToxCast, we find that 2D graphs play the most important role. These four datasets are related to toxicity (or side effects). Although it is a very complex biological issue to explain, such properties can still be partially deduced from certain functional groups patterns contained in 2D graphs. Actually, medicinal chemists have developed such a database to provide them with necessary alerts of potential side effects in drug design \cite{Baell:2010ns}.
\item BBBP, which measures blood-brain barrier permeability, is mostly dominated by the following properties: liposolubility/water-solubility, molecular weight, and interaction between molecules and transporter proteins. Similarly, these properties can also be inferred from 2D topology, such as molecules with too many hydrogen bond acceptors/donors are unlikely to break the blood-brain barrier due to poor liposolubility \cite{Suckling:1986bb}.
\item On BACE and MUV we see 2D graphs and SMILES strings contribute most. These two datasets are about predicting protein-ligand binding activities, which are theoretically relevant to 3D conformations. However, it is still an open question that whether the conformation sampling methods can produce conformations that resemble bioactive conformations, which provide the key information for protein-ligand binding. Nevertheless, in each of these tasks, the target protein is fixed so that bioactivity can be partially deduced from 2D structures, which is supported by the success of fragment-based Quantitive Structure-Activity Relationship (QSAR) models \cite{Manoharan:2010qs}.
\item Due to the complicated pathogenetic mechanisms, it is hard to draw an explanation to why attention weights of fingerprints outweigh the other three features in the HIV task. Given that the HIV dataset is the largest one (over 40,000 molecules per task), one possible explanation of this phenomenon is that we use a high-dimensional fingerprint representations (1024 bits).
\end{itemize}

Concerning the difference between three 2D-based features (namely 2D topological graphs, fingerprints, and SMILES strings), we make the following findings, which we hope could serve as guidelines for future research on molecular representation learning:
\begin{itemize}
\item 2D graph representations can encode local information explicitly by resembling chemical structures. Besides, graph-based neural networks can capture long-range local chemical environment through message passing. For example, with molecular graphs, it is more convenient to identify which part of the molecule serves as a scaffold.
\item In principle, SMILES strings contain all 2D information of certain molecules, but with atoms and bonds represented in ASCII characters, neural networks may have difficulty in distilling semantic meanings of chemical structures in a numerical way.
\item Fingerprint representations are based on local structures and thus such features may be less effective in circumstances where long-range effects induced by topologically distant functional groups predominate, which accounts for relatively small attention weights of fingerprints in \cref{fig:attention}.
\end{itemize}

\subsection{More Experiments on Molecular Property Regression}

\begin{table}
    \centering
    \caption{Results for eight molecule quantum property regression tasks in terms of Mean Absolute Error (MAE, \(\downarrow\)). The highest performance is highlighted in \textbf{bold}.}
    \label{tab:quantum-regression}
    \begin{tabular}{cccccccccc}
    \toprule
    Target & \(\mu\) & \(\alpha\) & \(\epsilon_\text{HOMO}\) & \(\epsilon_\text{LUMO}\) & \(\epsilon_\text{gap}\) & \(U_0\) & \(U\) & \(\left<R^2\right>\) \\
    Unit & D & Bohr\textsuperscript{3} & meV & meV & meV & meV & meV & Bohr\textsuperscript{3}\\
    \midrule
    SchNet-NP & 0.4604 & 0.3251 & 95.9740 & 78.5870 & 136.4720 & 98.1240 & 100.1650 & 24.3277 \\
    \themodel-NP & 0.3767 & 0.2439 & 73.0625 & 69.8780 & 102.2332 & 77.4708 & 92.8562 & 17.5842 \\
    \midrule
    GraphMVP & 0.3726  & 0.4390  & 75.3750  & 72.3820  & 104.8370  & 278.8900  & 325.8021  & 22.6433  \\
    3D Infomax & 0.3644  & 0.4190  & 72.0558  & 67.6203  & 99.4032  & 207.2148  & 219.5415  & 20.3934 \\
    \themodel & \textbf{0.3618} & \textbf{0.2236} & \textbf{71.5120} & \textbf{58.5890} & \textbf{97.7440} & \textbf{64.3550} & \textbf{66.3958} & \textbf{15.5571} \\
    \bottomrule
    \end{tabular}
\end{table}

To demonstrate that the conformations generated by RDKit are helpful, we further conduct an experiment on quantum property regression on the QM9 dataset \cite{Ramakrishnan:2014ij}, where 3D conformations generated by RDKit are used for the fine-tuning datasets. This task is known to be closely related to 3D structures.
\cref{tab:quantum-regression} presents the performance comparison of MEMO with two non-pretraining (supervised) baselines SchNet and \themodel (denoted by SchNet-NP and \themodel-NP) and two state-of-the-art pretraining baselines GraphMVP \cite{Liu:2022vr} and 3D Infomax \cite{Stark:2021ug}.

It is seen that our \themodel model achieves the best performance on all datasets.
GraphMVP that consider only 2D structures during fine-tuning even result in negative transfer on some datasets. Our \themodel, on the contrary, achieves better performance than the supervised baseline, underscoring the value of leveraging 3D structures (as well as other sources of 2D information) during fine-tuning.

We also perform experiments on non-quantum property regression tasks. Our proposed \themodel also obtains promising improvements compared to the current state-of-the-art baselines. Please refer to \cref{supp:property-regression} for performance comparison and analysis.

\subsection{Ablation Studies}
\begin{figure}[b]
	\centering
	\includegraphics[width=\linewidth,bb=0 0 601 156]{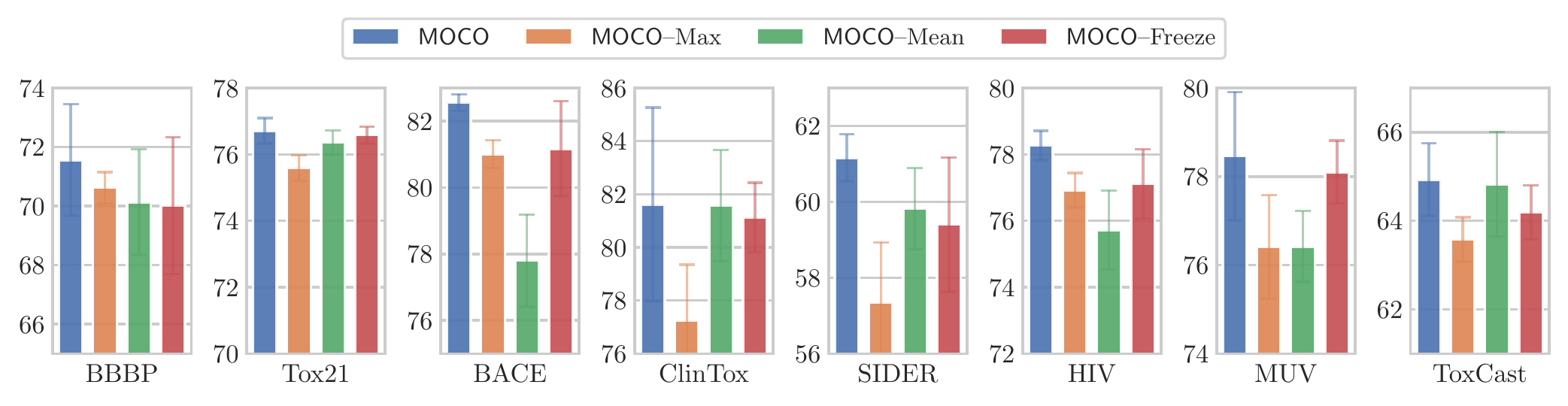}
	\caption{Ablation studies on representation aggregation and the fine-tuning strategy.}
	\label{fig:ablation}
\end{figure}

Finally, we conduct ablation studies on the representation aggregation module and the fine-tuning strategy.
We consider the following model variants for further inspection.
Except the modifications in specific modules, other implementations remain the same as previously described.
\begin{itemize}
	\item \textbf{\themodel --Max} removes the attention network in the representation aggregation module in \cref{eq:attention-aggregation} and simply uses max pooling to combine view embeddings.
	\item \textbf{\themodel --Mean} modifies representation aggregation by taking average over view embeddings.
	\item \textbf{\themodel --Freeze} does not fine-tune the representation aggregation module but instead uses the frozen weights of the pretrained model.
\end{itemize}

We report the performance of model variants in \cref{fig:ablation}.
It is seen that all three variants achieve downgraded performance, which empirically rationalizes the design choice of our molecular pretraining framework with complementary featurizations.
Specifically, the performance of \themodel --Max and \themodel --Mean without attention aggregation mechanisms of multiple featurizations is inferior to that of \themodel, demonstrating the necessity of adaptively combining information from multiple featurizations.
In addition, \themodel --Freeze occasionally obtains better performance than the two other variants, which indicates that our proposed attention network is able to select information from different views.
It does not, however, fine-tune the contribution of featurizations with downstream datasets, where the optimal combination might differ, resulting in performance deterioration.

Moreover, we conduct ablation studies on models that include only three view representations, where the results can be found in \cref{supp:more-ablation}. Results demonstrate the necessity of comprehensively leveraging four views in the proposed \themodel model.

\section{Related Work}
\label{sec:recent-work}

Traditional methods \cite{Carhart:1985ap,Nilakantan:1987tt,Rogers:2010fp} represent molecular structures with fingerprints. Some prior studies \cite{Svetnik:2004ab,Meyer:2019ld,Wu:2018dv} employ tree-based machine leaning models such as random forests \cite{Breiman:2001rf} and XGBoost \cite{Chen:2016ga} on fingerprints to predict the properties of molecules.
With the development of deep learning, neural approaches have been dominating the field given their strong representation ability.
One line of work \cite{Wang:2019hp,Chithrananda:2020eo} leverages language modeling techniques such as BERT \cite{Devlin:2019uk} to learn molecular representations based on SMILES strings \cite{Weininger:1988sm}.
However, some argue that sequence-based representations cannot fully capture substructure information and propose to leverage Graph Neural Networks (GNNs), which model molecules as graphs with atoms as nodes and bonds as edges \cite{Gilmer:2017tl,Liu:2019uy,Ying:2021ug}.
Despite the prosperous progress, they only model 2D topological structures of molecules, without considering the 3D coordinates of atoms that are known to determine certain chemical and physical functionalities of molecules.
To address this deficiency, recent work further explicitly considers such 3D geometry and designs equivariant networks to obtain the representations \cite{Schutt:2017wh,Klicpera:2020vw,Satorras:2021tz,Fuchs:2020wj,Schutt:2021vm,Du:2021ci,Liu:2021hq,Gasteiger:2021uf,Batzner:2021to,Brandstetter:2022wl,Xu:2021uj}.

Even though molecular representation learning techniques have been extensively investigated, there are very few labeled datasets available for studying the molecular properties of interest (e.g., drug-likeness or quantum properties).
On the other hand, there are abundant unannotated molecules available, which motivates researchers to study pretraining techniques that learn the model weights in a self-supervised manner and transfer the knowledge to downstream datasets with limited annotations via fine-tuning.
A series of pretraining frameworks on 2D molecular graph representations have been developed so far \cite{Rong:2020vk,Hu:2020uz,Zhang:2021wj,Wang:2022gr,Li:2020fo,Xia:2022jw}.
Recent work GEM \cite{Fang:2022et} studies large-scale pretraining for 3D geometry representations.
Additionally, researchers also study to supplement 2D-graph-based pretraining with 3D conformation information \cite{Yang:2021wg,Liu:2022vr,Stark:2021ug}.

A succinct comparison of our work with other representative methods is provided in \cref{tab:comparison-baseline}.
Compared to the above studies, our proposed \themodel is the only model that can \emph{adaptively} leverage multiple featurizations for both pretraining and fine-tuning stages.

\begin{table}
	\centering
	\rowcolors{2}{white}{lightgray!10}
	\caption{Comparing \themodel with representative self-supervised methods on molecular pretraining.}
	\begin{tabular}{l*{8}{c}}
	\toprule
	& \multicolumn{4}{c}{Pretraining} & \multicolumn{4}{c}{Fine-tuning} \\
	\cmidrule(lr){2-5} \cmidrule(lr){6-9}
	\rowcolor{white} \multirow{-2.5}{*}{Method} & 2D & 3D & Fingerprint & SMILES & 2D & 3D & Fingerprint & SMILES \\
	\midrule
	SMILES-BERT \cite{Wang:2019hp} & & & & \cmark & & & & \cmark \\
	ChemBERTa \cite{Chithrananda:2020eo} & & & & \cmark & & & & \cmark \\
%	GraphSAGE \cite{Hamilton:2017tp} & \cmark & & & & \cmark & & & \\
	AttrMask, ContexPred \cite{Hu:2020uz} & \cmark & & & & \cmark & & & \\
%	GPT-GNN \cite{Hu:2020vh} & \cmark & & & & \cmark & & & \\
%	InfoGraph \cite{Sun:2020vi} & \cmark & & & & \cmark & & & \\
	GraphCL \cite{You:2020ut} & \cmark & & & & \cmark & & & \\
%	JOAO \cite{You:2021wl} & \cmark & & & & \cmark & & &  \\
	GraphLoG \cite{Xu:2021tv} & \cmark & & & & \cmark & & & \\
	GROVER \cite{Rong:2020vk} & \cmark & & & & \cmark & & & \\
	GEM \cite{Fang:2022et} & & \cmark & & & & \cmark & & \\
	3D Infomax \cite{Stark:2021ug} & \cmark & \cmark & & & \cmark & & &  \\
	GraphMVP \cite{Liu:2022vr} & \cmark & \cmark & & & \cmark & & &  \\
	\themodel (Ours) & \cmark & \cmark & \cmark & \cmark & \cmark & \cmark & \cmark & \cmark \\
	\bottomrule
	\end{tabular}
	\label{tab:comparison-baseline}
\end{table}

\section{Conclusions and Discussions}
\label{sec:conclusion}

This paper examines different featurizations for molecular data and highlights the importance of incorporating multiple featurizations during both pretraining and fine-tuning.
Then, we develop a novel pretraining framework \themodel with complementary featurizations for molecular data, which is able to adaptively distill information from each featurization and allows interpretability from the learned model weights.
Extensive experiments on a wide range of property prediction benchmarks show that \themodel consistently outperforms existing baselines without negative transfer.

The study of featurization techniques for molecular machine learning in general remains widely open.
We would like to acknowledge that the relative utility of various featurizations for different molecular predictive tasks could be usefully explored in further work.
Moreover, more future research should be undertaken to specifically analyze the relationship between several featurizations, the representation ability of corresponding neural architectures, as well as the task-featurization correlation.

\bibliographystyle{unsrtnat}
\bibliography{reference}

\clearpage
\appendix
\appendixprefix

\section{Implementation of View Encoders}
\label{supp:view-encoders}

In this section, we introduce the detailed implementation of the four view encoders.
We denote the representation for node (atom) \(v_i\) as \(\bm{h}_i\) and the representation at the graph (molecule) level as \(\bm{z}\).
As each encoder is independent to each other, we omit the superscript representing the specific view \(m \in \{\text{2D}, \text{3D}, \text{FP}, \text{SM}\}\) is clear for notation simplicity.
Also, for clarity, when the context is clear, we omit the subscript \(j\) that indexes the molecule.

\paragraph{Embedding 2D graphs.}
Graph Isomorphism Network (GIN) \cite{Xu:2019ty} is a simple and effective model to learn discriminative graph representations, which is proved to have the same representational power as the Weisfeiler-Lehman test \cite{Weisfeiler:1968wl}.
Since GIN has been widely adopted for 2D graph representation learning \cite{Hu:2020uz,You:2021wl,You:2020ut}, we leverage a GIN model to obtain the representations for the 2D molecular graphs.
Recall that each molecule is represented as \(\mathcal{G} = (\bm{A}, \bm{X}, \tens{E})\), where \(\bm{A}\) is the adjacency matrix, \(\bm{X}\) and \(\tens{E}\) are features for atoms and bonds respectively.
The layer-wise propagation rule of GIN can be written as:
\begin{equation}
	\bm{h}_i^{(k + 1)} = f_{\text{atom}}^{(k+1)}\left(
	\bm{h}_i^{(k)} + \sum_{j \in\mathcal{N}(i)}
	\left(\bm{h}_j^{(k)} + f_{\text{bond}}^{(k+1)}(\etens{E}_{ij}))
	\right)\right),
\end{equation}
where the input features \(\bm{h}^{(0)}_i = \bm{x}_i\), \(\mathcal{N}(i)\) is the neighborhood set of atom \(v_i\), and \(f_{\text{atom}}, f_{\text{bond}}\) are two MultiLayer Perceptron (MLP) layers for transforming atoms and bonds features, respectively.
By stacking \(K\) layers, we can incorporate $K$-hop neighborhood information into each center atom in the molecular graph.
Then, we take the output of the last layer as the atom representations and further use the mean pooling to get the graph-level molecular representation:
\begin{equation}
\bm{z}^\text{2D} = \frac{1}{N}\sum_{i\in \mathcal{V}}\bm{h}_i^{(K)}.
\end{equation}

\paragraph{Embedding 3D graphs.}
Following GraphMVP \cite{Liu:2022vr}, we use the SchNet \cite{Schutt:2017wh} as the encoder for the 3D geometry graphs.
SchNet models message passing in the 3D space as continuous-filter convolutions, which is composed of a series of hidden layers, given as follows:
\begin{equation}
    \bm{h}_i^{(k + 1)} = f_\text{MLP}\left(\sum_{j=1}^N f_\text{FG}(\bm{h}_j^{(t)}, \bm{r}_i, \bm{r}_j)\right) + \bm{h}_i^{(t)},\\
\end{equation}
where the input \(\bm{h}^{(0)}_i = \bm{a}_i\) is an embedding dependent on the type of atom \(v_i\), \(f_\text{FG}(\cdot)\) denotes the filter-generating network. 
To ensure rotational invariance of a predicted property, the message passing function is restricted to depend only on rotationally invariant inputs such as distances, which satisfying the energy properties of rotational equivariance by construction.
Moreover, SchNet adopts radial basis functions to avoid highly correlated filters. The filter-generating network is defined as follow:
\begin{equation}
	f_\text{FG}(\bm{x}_j, \bm{r}_i, \bm{r}_j) = \bm{x}_j\cdot e_k(\bm{r}_i - \bm{r}_j) = \bm{x}_j \cdot \exp (-\gamma \|\|\bm{r}_i - \bm{r}_j\|_2 - \mu \|_2^2).
\end{equation}
Similarly, for non-quantum properties prediction concerned in this work, we take the average of the node representations as the 3D molecular embedding:
\begin{equation}
	\bm{z}^\text{3D} = \frac{1}{N}\sum_{i\in \mathcal{V}}\bm{h}_i^{(K)},
\end{equation}
where $K$ is the number of hidden layers.

\paragraph{Embedding fingerprints.}
Due to the discrete and extremely sparse nature of fingerprint vectors, we first transform all \(F\) binary feature fields into a dense embedding matrix \(\bm{F}' \in \mathbb{R}^{F \times D_\text{F}}\) via embedding lookup.
Then, we introduce a positional embedding matrix \(\bm{P} \in \mathbb{R}^{F \times D_\text{F}}\) to capture the positional relationship among bits in the fingerprint vector, which is defined as:
\begin{align}
	\bm{P}_{p, 2i} &= \sin(p/10000^{2i/D_F}),\\
	\bm{P}_{p, 2i+1} &= \cos(p/10000^{2i/D_F}),
\end{align}
where \(p\) denotes the corresponding bit position and \(i\) is corresponds to the \(i\)-th embedding dimension.
The positional embedding matrix will be added to the transformed embedding matrix:
\begin{equation}
	\bm{F} = \bm{F}' + \bm{P}.
\end{equation}
Thereafter, we use a multihead Transformer \cite{Vaswani:2017ul} to model the interaction among those feature fields.
Specifically, we first transform each feature into a new embedding space as:
\begin{align}
	\bm{Q}^{(h)} & = \bm{F}\bm{W}_\text{Q}^{(h)}, \\
	\bm{K}^{(h)} & = \bm{F}\bm{W}_\text{K}^{(h)}, \\
	\bm{V}^{(h)} & = \bm{F}\bm{W}_\text{V}^{(h)},
\end{align}
where the three linear transformation matrices \(\bm{W}_\text{Q}^{(h)}, \bm{W}_\text{K}^{(h)}, \bm{W}_\text{V}^{(h)} \in \mathbb{R}^{D_\text{F} \times \nicefrac{D}{H}}\) parameterize the query, key, and value transformations for the \(h\)-th attention head, respectively.
Following that, we compute the attention scores among all feature pairs and then linearly combine the value matrix from all \(H\) attention heads:
\begin{align}
	\bm{W}_\text{A}^{(h)} & = \operatorname{softmax}\left(\frac{\bm{Q}^{(h)}(\bm{K}^{(h)})^\top}{\sqrt{D_\text{H}}}\right), \\
	\widehat{\bm{Z}} & = \left[\bm{W}_\text{A}^{(1)}\bm{V}^{(1)} \,;\, \bm{W}_\text{A}^{(2)}\bm{V}^{(2)} \,;\, \dots \,;\, \bm{W}_\text{A}^{(H)}\bm{V}^{(H)} \right],
\end{align}
Finally, we perform sum pooling on the resulting embedding matrix \(\widehat{\bm{Z}} \in \mathbb{R}^{F \times D_\text{F}}\) and use a linear model \(f_\text{LIN}\) to obtain the final fingerprint embedding \(\bm{z}^\text{FP} \in \mathbb{R}^{D}\):
\begin{equation}
	\bm{z}^\text{FP} = f_\text{LIN} \left(\sum_{d=1}^{D_\text{F}} \widehat{\bm{Z}}_{d}\right).
\end{equation}

\paragraph{Embedding SMILES strings.}
Given ASCII-encoded SMILES strings, we first tokenize them with the Byte-Pair Encoder (BPE) tokenizer \cite{Gage:1994}, which strikes a balance among character- and word-level representations and allows to handle large vocabularies in molecular corpora.
Specifically, BPE finds the best word segmentation by iteratively and greedily merging frequent pairs of characters. In our implementation, we use a max vocabulary size of 52K tokens for both pretraining and downstream datasets.

After tokenization, we first pretrain a RoBERTa \cite{Liu:2019dd} model on the pretraining dataset with the masking language model as the sole training objective, as SMILES strings do not possess sequential relationships.
To be specific, $15\%$ tokens in a SMILES string are randomly selected and replaced with a special token \token{MASK}.
We also insert a special token \token{CLS} to each string to represent the whole molecule.
The training objective function is to independently predict the original tokens given the output on masked tokens.
Finally, the representation of the \token{CLS} token is regarded as the molecular embedding.

After pretraining the RoBERTa backbone, we freeze its parameters and leverage an additional MLP layer on top of each molecular embedding to obtain the final representation for each SMILES string.
This strategy improves memory efficiency and thus enables larger batch sizes for our contrastive pretraining framework.

\section{Dataset Description}
\label{supp:dataset}

In this section, we briefly introduce the datasets used for pretraining and fine-tuning, as well as details of dataset prepossessing.
Basic dataset statistics is summarized in \cref{tab:dataset}.

\begin{table}
	\centering
	\caption{Statistics of datasets used in experiments. The first section describes the datasets with 3D information which is used for pre-training; the later two sections describe datasets for fine-tuning.}
    \begin{tabular}{ccccccc}
    \toprule
    & Dataset & \#Molecules & Avg. \#atoms & Avg. \#bonds & \#Tasks & Avg. degree \\
    \midrule
    & GEOM-Drug & 304,466 & 44.40 & 46.40  & --- & 2.09 \\
    \midrule
    \parbox[t]{0mm}{\multirow{8}{*}{\rotatebox[origin=c]{90}{Classification}}} & BBBP  & 2,039 & 24.06 & 25.95 & 1     & 2.16  \\
     & Tox21 & 7,831 & 18.57 & 19.29   & 12    & 2.08  \\
     & ToxCast & 8,576 & 18.78 & 19.26 & 617   & 2.05  \\
     & SIDER & 1,427 & 33.64 & 35.36 & 27    & 2.10  \\
     & ClinTox & 1,477 & 26.16 & 27.88 & 2     & 2.13  \\
     & MUV   & 93,087 & 24.23 & 26.28 & 17    & 2.17  \\
     & HIV   & 41,127 & 25.51 & 27.47 & 1     & 2.15  \\
     & BACE  & 1,513 & 34.09 & 36.86 & 1     & 2.16  \\
    \midrule
    \parbox[t]{0mm}{\multirow{5}{*}{\rotatebox[origin=c]{90}{Regression}}} & ESOL  & 1,128 & 13.30 & 13.69 & 1 & 2.06  \\
     & Lipophilicity  & 4,200 & 27.04 & 29.50 & 1     & 2.18  \\
     & Malaria & 9,999 & 30.36 & 33.20 & 1     & 2.19  \\
     & CEP   & 29,978 & 27.66 & 33.39 & 1     & 2.41  \\
     & QM9   & 130,831 & 18.03 & 18.65 & 8     & 2.07  \\
    \bottomrule
    \end{tabular}
	\label{tab:dataset}
\end{table}

\subsection{Pretraining Datasets}
We choose GEOM-Drugs\footnote{\url{https://github.com/learningmatter-mit/geom}} \cite{Axelrod:2022da} as the pre-training dataset, which contains high-quality conformers for 304,466 mid-sized organic molecules with experimental data.
The conformer information in GEOM-Drugs is generated using the CREST \cite{Grimme:2019ec} program, which provides reliable and accurate structure generation.
Note that atoms usually have multiple conformations resulting in potentially different chemical properties.
In our work, we focus on the conformations of the lowest energy, as they are more likely to occur naturally \cite{Stark:2021ug,Liu:2022vr}.
Moreover, since the original full set is large (317K molecules with over 9M conformations), we follow GraphMVP \cite{Liu:2022vr} to sample a subset of 50K molecules, each with its top-5 conformations, for pretraining. We use the same random seeds to ensure dataas with GraphMVP.

\subsection{Fine-Tuning Datasets}
For fine-tuning, we use 12 datasets collected from MoleculeNet\footnote{\url{https://github.com/deepchem/deepchem}}\cite{Wu:2018dv}, ChEMBL \cite{Gaulton:2011ch}, and CEP \cite{Hachmann:2011ce}, which target on different properties and distinct tasks. These properties can be divided into four main categories: physical chemistry, biophysics, physiology, and quantum properties.

\paragraph{Physical chemistry.}
ESOL \cite{Delaney:2004es} consists of water solubility data recording whether molecules are water-soluble.
The Lipophilicity dataset is a subset of ChEMBL \cite{Gaulton:2011ch} measuring the molecule octanol/water distribution coefficient.
The CEP dataset is a subset of the Havard Clean Energy Project (CEP) \cite{Hachmann:2011ce}, which estimates the organic photovoltaic efficiency.

\paragraph{Biophysics.}
The HIV dataset \cite{AIDS:as} is introduced by Drug Therapeutics Program (DTP) AIDS Antiviral Screen, which tests the molecular ability to inhibit HIV replication.
The Maximum Unbiased Validation (MUV) group \cite{Rohrer:2009mu} is another benchmark dataset selected from PubChem BioAssay by applying a refined nearest neighbor analysis.
The BACE dataset provides qualitative binding results for a set of inhibitors of human \(\beta\)-secretase 1 (BACE-1).
The Malaria dataset \cite{Gamo:2010wo} assesses the drug efficacy in inhibiting parasites that cause malaria.

\paragraph{Physiology.}
The Blood–brain barrier penetration (BBBP) dataset \cite{Martins:2012ba} models the barrier permeability of molecules targeting central nervous system.
Tox21 \cite{Tox21:2014cl}, ToxCast \cite{Richard:2016tc}, and ClinTox \cite{Gayvert:2016ad} are all related to the toxicity of molecular compounds.
The Side Effect Resource (SIDER) \cite{Kuhn:2016sd} is a dataset measuring the adverse drug reactions of 27 system organ classes of marketed drugs.

\paragraph{Quantum properties.}
The QM9 dataset \cite{Ramakrishnan:2014ij} consists of 134K stable small organic molecules made up of four kinds of heavy atoms (C, O, F, N) and hydrogen (H), and each atom is associated with 12 quantum mechanical properties as regression targets, which are closely related to geometries minimal in energy.
In our experiments, we delete 3,054 uncharacterized molecules which failed the geometry consistency check \cite{Ramakrishnan:2014ij} and adopt the remaining 130,831 molecules for the downstream task with 8 quantum-related property regression targets. Specifically, the property $\mu$ represents the dipole moment of the molecule, the property $\alpha$ represents the corresponding isotropic polarizability, and the $\left<R^2\right>$ denotes the electronic spatial extent. The property $\epsilon_{\text{HOMO}}, \epsilon_{\text{LUMO}}$, and $\epsilon_{\text{gap}}$ represent the energy of HOMO, the energy of LUMO, and the energy gap of the former two energies respectively. The property $U_0$ and $U$ denote the internal energy of the molecule at 0K and 298.15K, respectively.

\paragraph{A note on overlapping molecules.}
We find that only 14,087 ($\approx 28.2\%$) molecules in the MUV dataset overlap with those in GEOM-Drugs and there is no overlapping molecules in the other fine-tuning datasets.
Considering that \emph{no} labels is used during pretraining, for those overlapped molecules, we are evaluating the in-domain performance (unsupervised learning performance), while for those non-overlapped ones, we are evaluating the transfer learning performance.

\subsection{Dataset Preprocessing}

For classification tasks, we leverage atom types and chirality tags as atom attributes, while the type and direction of the bond are corresponding bond attributes. Both the atom and bond attributes are expressed in the form of discrete indices without further embedding.
For regression tasks, we first transform discrete atom and bond attributes through learnable embedding lookup layers following OGB \cite{Hu:2020wv}.
Since molecules in the fine-tuning datasets do not have 3D information available, we use ETKDG \cite{Riniker:2015bi} in RDkit \cite{Landrum:2022rd} to generate molecular conformations.

\paragraph{Constructing fingerprints.}
For both pretraining and fine-tuning datasets, we use RDkit to generate molecular fingerprints.
Morgan fingerprints \cite{Morgan:1965tg,Glem:2006cf} encode molecules in fixed-length binary strings, with bits indicating presence or absence of specific substructures.
The algorithm assigns an initial identifier to each non-hydrogen atom according to a set of atomic invariants, iteratively updates the identifiers among neighborhood atoms within certain hops, and encodes the identifiers using a hash function.
After hashing all of these identifiers into a fixed-length binary string, the representation provides information on topological characteristics of the molecule.

In our implementation, we set the diameter of neighborhood to 2, the length of fingerprints to 1024, and follow the default configuration of ECFP4 \cite{Rogers:2010fp}, which uses the following connectivity invariants to construct the initial node features:
\begin{itemize}[nosep]
	\item The atomic number
	\item The number of heavy (non-hydrogen) neighbor atoms
	\item The number of attached hydrogens
	\item The formal charge
	\item Atom isotopes
	\item Whether the atom is part of at least one ring
\end{itemize}

\section{Implementation Details}
\label{supp:implementation}

\subsection{Computing Infrastructures}
\paragraph{Software infrastructures.} 
The proposed \themodel is implemented in Python 3.7, with the following supporting libraries: PyTorch 1.10.2 \cite{Paszke:2019vf}, PyG 2.0.3 \cite{Fey:2019wv}, RDKit 2022.03.1 \cite{Landrum:2022rd}, OGB 1.3.3 \cite{Hu:2020wv} and HuggingFace's Transformers 4.17.0 \cite{Wolf:2020iu}.

\paragraph{Hardware infrastructures.} 
We conduct all experiments on a computer server with 8 NVIDIA GeForce RTX 3090 GPUs (with 24GB memory each) and 256 AMD EPYC 7742 CPUs.

\subsection{Code Availability}
For the GIN backbone model for 2D topology graphs, we adopt its official implementation \cite{Hu:2020uz}.
For SchNet and all the other backbone models, we use the source codes provided by GraphMVP \cite{Liu:2022vr} and follow the settings in its original paper for fair comparison.

\subsection{Hyperparameter Specifications}
All model parameters are initialized with the Glorot initialization \cite{Glorot:2010uc} and trained using the Adam optimizer \cite{Kingma:2015us}.

For hyperpameters of the GIN model and the SchNet model, we keep them in line with the baseline GraphMVP \cite{Liu:2022vr}. Specifically, for the GIN model, we set the number of convolutional layers to 5 and hidden dimension to 300;
for the SchNet model, we set the hidden dimension and the number of filters in continuous-filter convolution to 128. The interatomic distances are measured with 50 radial basis functions. We stack 6 interaction layers in the SchNet architecture.
For the RoBERTa model, we use the default configuration. Particularly, the hidden dimension is set to 768 and the number of attention heads is set to 12, and the dropout ratio between hidden layers is set to 0.1.
In addition, we add a MLP layer with the hidden dimension of 300 as the prediction head.  

For the fingerprint encoder, we set the bit embedding dimension and the hidden dimension to 64 and 300, respectively. Moreover, we adopt a one-layer Transformer model with 8 attention heads.
For the representation aggregation module, similarly, we set the hidden dimension of all the trainable parameters to 300.
For all activation functions, we stick to \(\operatorname{ReLU}(\cdot) = \max(0, \cdot)\).
For the temperature parameter \(\tau\) in the contrastive objective, we tune it from \(\{0.1, 0.3, 0.5\}\).
All these hyperparameters except for those in GIN and SchNet backbones are selected based on grid search on the validation set.

Moreover, we empirically find that the dropout ratio \cite{Srivastava:2014cg} and the learning rate are two important hyperparameters in our model.
Due to different training dynamics of different view encoders, we do a hyperparameter search of the learning rates and dropout ratio for each encoder from $\{10^{-3},10^{-4},\dots,10^{-7}\}$ and $\{0, 0.3, 0.5\}$, respectively.
We would like to emphasize that we do not tune hyperparameters for different downstream datasets.

\subsection{Experimental Configurations}

To obtain the SMILES embeddings, the RoBERTa backbone is firstly pretrained for 4 epochs with all of the available molecules in the GEOM-Drugs dataset.
Then, during the pretraining stage, we set train the whole model for 100 epochs with a batch size of 256 and we ensure that the model converges.

In the fine-tuning stage, for fair comparison, we follow the scaffold split adopted by GraphMVP \cite{Liu:2022vr} and repeat the experiments for three times.
Specifically, we fine-tune our model on downstream tasks with three different random splits and we report the averaged test performance with the best models obtained on the validation set for all the methods.

\subsection{Computational Complexity Analysis}
In \themodel, calculating featurizations (i.e. fingerprints, SMILES strings, and 2D/3D graphs) is very efficient: computing fingerprints based on SMILES strings can be done in an instant. For 3D structures, there are two types of conformations involved: for pretraining, we directly use the DFT-calculated conformations shipped with the dataset; for fine-tuning, as no such conformation is available, we sample conformation with RDKit which is efficient ($\approx$0.01 sec/molecule). Note that it is costly to use DFT to optimize molecular conformation. An example is given by the OpenCatalyst project\footnote{\url{https://opencatalystproject.org/challenge.html}}, which reports that a standard relaxation using DFT takes 8 to 10 hours.
In terms of pretraining time, our \themodel takes slightly longer than GraphMVP's. To be specific, our pretraining on GEOM-Drugs takes $\approx$4h17m to finish whereas the baseline GraphMVP takes $\approx$4h03m.

\section{Additional Experiments}
\label{supp:additional-experiments}

In this section, we further perform experiments on non-quantum property regression tasks. We also include additional experiments with ablated models and model variants.

\subsection{More Experiments on Molecular Property Regression Tasks}
\label{supp:property-regression}

We further conduct experiments on four additional regression tasks for molecular property prediction, where the results are presented in \cref{tab:regression}.
It can be clearly seen from the table that our \themodel considerably improves the performance of baselines on three datasets and achieves similar performance to the baseline approaches on the Lipophilicity dataset, which once again verifies the effectiveness of our framework and demonstrates the importance of integrating different molecular featurization techniques.

\begin{table}
  \centering
  \caption{Additional results on four molecular property regression tasks in terms of Root-Mean-Square Error (RMSE). The lowest prediction error is highlighted in \textbf{boldface}.}
    \begin{tabular}{lccccc}
    \toprule
    Pretraining & ESOL & Lipophilicity & Malaria & CEP & Avg. \\
    \midrule
    ---   & 1.364{\tiny±0.016} & 0.736{\tiny±0.006} & 1.122{\tiny±0.011} & 1.380{\tiny±0.033} & 1.15051 \\
    \midrule
    AttrMask & 1.112{\tiny±0.048} & 0.730{\tiny±0.004} & 1.119{\tiny±0.014} & 1.256{\tiny±0.000} & 1.05419 \\
	ContextPred & 1.196{\tiny±0.037} & \textbf{0.702{\tiny±0.020}} & 1.101{\tiny±0.015} & 1.243{\tiny±0.025} & 1.06059 \\
	JOAO & 1.120{\tiny±0.019} & 0.708{\tiny±0.007} & 1.145{\tiny±0.010} & 1.293{\tiny±0.003} & 1.06631 \\
	GraphMVP & 1.091{\tiny±0.021} & 0.718{\tiny±0.016} & 1.114{\tiny±0.013} & 1.236{\tiny±0.023} & 1.03968 \\
    \midrule
    \themodel  & \textbf{0.984{\tiny±0.034}} & 0.707{\tiny±0.001} & \textbf{1.093{\tiny±0.009}} & \textbf{1.101{\tiny±0.007}} & \textbf{0.97125} \\
    \bottomrule
    \end{tabular}
	\label{tab:regression}
\end{table}

\subsection{Ablation Studies on Multiple Views}
\label{supp:more-ablation}

\begin{figure}
	\centering
	\includegraphics[width=\linewidth]{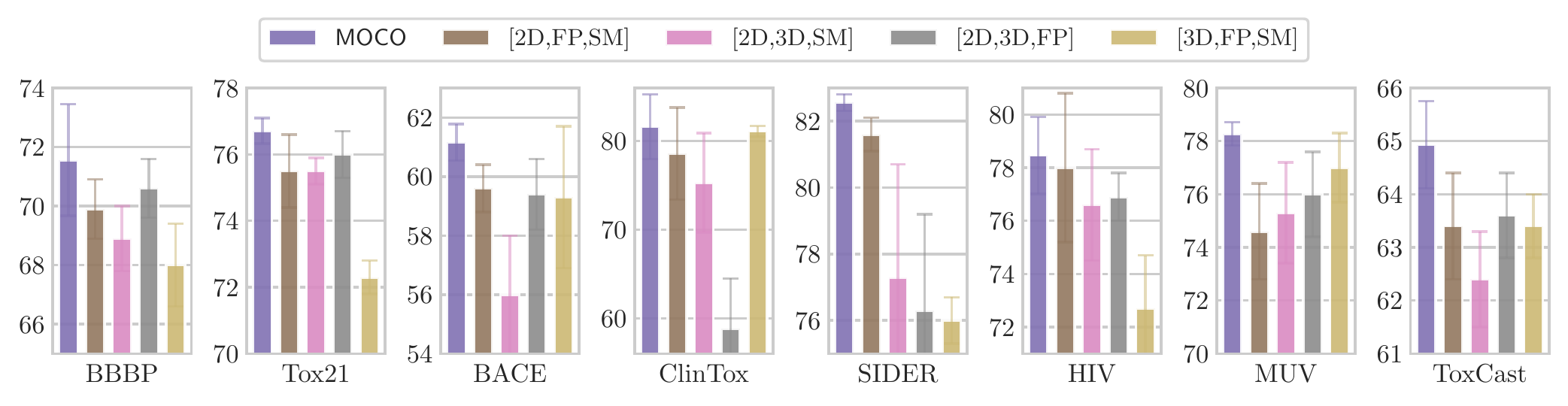}
	\caption{Ablation studies on four featurization techniques used in \themodel.}
	\label{fig:view-ablation}
\end{figure}

To further verify the necessity of incorporating all four views (2D, 3D, FP and SM), we further compare our \themodel model with four featurization techniques with its ablated counterparts. \cref{fig:view-ablation} summarizes the model performance obtained with only three views.
From the table, it is observed that our \themodel that adaptively learns to optimize the combinations of four featurization techniques for different downstream tasks achieves the best performance for almost all datasets, which demonstrates the necessity of comprehensively considering four featurization.
We kindly note that our framework is general and flexible; it is thus not limited to specifically incorporate these four views.

\subsection{Experiments on Variant Fingerprint Encoders}

\begin{figure}
	\centering
	\includegraphics[width=\linewidth]{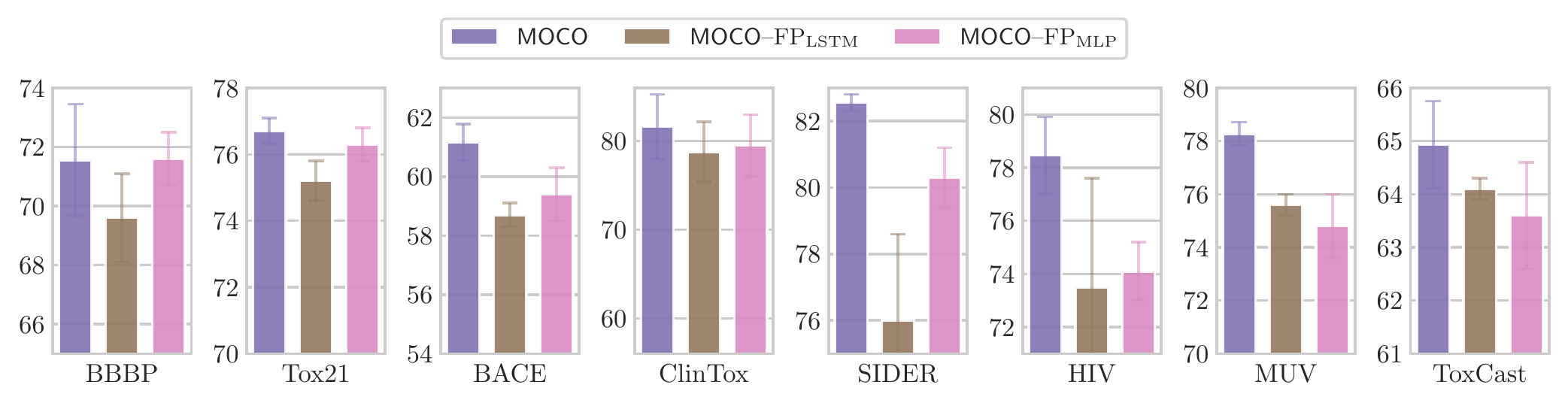}
	\caption{Results with two variant encoders for fingerprints: MLP and LSTM.}
	\label{fig:fp-variants}
\end{figure}

Since there is a lack of proper neural encoders for fingerprints, we propose an attention-based network to model interactions of feature fields in fingerprint vectors, which considers the discrete and extremely sparse nature of fingerprints.
To demonstrate its effectiveness, we leverage a simple MLP and a Long Short-Term Memory (LSTM) model as the encoder for the fingerprints respectively.
The performance comparison is summarized in \cref{fig:fp-variants}. It is clear that our proposed attention encoder that models feature interactions for fingerprints achieves the best performance.

\section{More Related Work}
\label{supp:related-work}

The following section provides a more broad literature review across the spectrum of self-supervised representation learning.

\subsection{Self-Supervised Representation Learning on Visual and Natural Language Data}

A SSL model trains itself by learning a part of the input from another through pretext tasks.
Depending on the pretext task, the existing SSL studies can be divided into three main categories.

Early SSL work studies \emph{predictive training} on pseudo-labels directly computed from the raw data.
In Computer Vision (CV) domains, typical pretext tasks include image inpainting \cite{Pathak:2016gb}, rearranging shuffled image patches \cite{Noroozi:2016hd}, colorizing grayscale images \cite{Zhang:2016fr,Larsson:2017vt}, recognizing geometric transformations \cite{Gidaris:2018wr}, and predicting cluster assignments \cite{Caron:2018ba}.
In Natural Language Processing (NLP), word2vec \cite{Mikolov:2013uz} popularizes this paradigm by proposing Continuous Bag-Of-Words (CBOW) and skip-gram models for predicting center and neighboring words, respectively.
Other exemplary work includes \citet{Kiros:2015uq} that predicts neighborhood sentences and BART \cite{Lewis:2020il} that recovers sentence permutation.

The second group of SSL is \emph{contrastive} learning, which seeks to maximize the agreement of embeddings in the latent space under stochastic data augmentations by contrasting positive and negative samples \cite{Jing:2021cf}.
It has revolutionized unsupervised representation learning in recent years \cite{vandenOord:2018ut,Bachman:2019wp,He:2020tu,Chen:2020wj,Caron:2020uv,Chen:2021ci,Gao:2021wf} and has been witnessed to perform on par with its supervised counterparts \cite{He:2020tu,Chen:2020wj}.
A key success to contrastive models is to leverage strong data augmentations that induce invariance irrelevant to properties of the end tasks \cite{Xiao:2021vt,Tian:2020vw,Purushwalkam:2020wm,vonKugelgen:2021wb}.

The third line of development focuses on \emph{generative modeling} of input data.
Its core idea is to randomly remove a portion of data and train the model to recover the removed content.
This so-called masked language modeling and its autoregressive counterparts are first pioneered in the NLP community \cite{Bengio:2003vh,Peters:2018jz,Devlin:2019uk,Liu:2019dd,Lan:2020tt,Radford:2018im,Radford:2019lm,Brown:2020tp} and have since gained increasing popularity in the CV domain \cite{Dosovitskiy:2021uc,He:2022hx,Wei:2021ev}.
Unlike contrastive learning, generative approaches do not rely on curated data augmentations.
It has been reported that they scale well and generalize to different downstream tasks \cite{Devlin:2019uk,Brown:2020tp,He:2022hx}.

\subsection{Graph Self-Supervised Representation Learning}
Analogous to the above studies on visual and natural language data, SSL approaches in the graph domain can also be organized into the same three categories.
Due to the rapid development of graph SSL, we only review the most representative studies in each group.
Readers may refer to recent surveys \cite{Wu:2022wg,Xie:2022uv,Liu:2022wc} for comprehensive reviews and \citet{Zhu:2021tu} for a benchmarking study.

Firstly, the pioneering \emph{predictive} model \citet{Hu:2020uz} explores four strategies at both node and graph levels, including masked attribute prediction, context prediction, supervised attribute prediction, and structural similarity prediction.
\citet{You:2020um} study three SSL tasks through a multi-task framework to enable predictive training of graph-structured data.
M3S \cite{Sun:2020wx} explores the use of cluster assignments \cite{Caron:2018ba} as pseudo-labels and proposes a self-training framework that incrementally adds high-confident nodes to the labeled dataset.

The second group of work studies \emph{generative} training.
GraphSAGE \cite{Hamilton:2017tp} performs the link prediction task to reconstruct the graph structure in a once-for-all manner, similar to graph autoencoders \cite{Kipf:2016ul}.
GPT-GNN \cite{Hu:2020vh} proposes to perform node and edge reconstruction iteratively.

Lastly, along the line of graph \emph{contrastive} learning, some investigate contrasting modes for graph data, typical work of which includes cross-scale contrasting \cite{Velickovic:2019tu,Hassani:2020un}, same-scale contrasting \cite{Sun:2020vi,Zhu:2020vf,You:2020ut}, and hierarchical contrasting \cite{Xu:2021tv,Lin:2021vt}.
Another line of work investigates data augmentations.
GraphCL \cite{You:2020ut} proposes four heuristic augmentation schemes including edge dropping, node dropping, attribute masking, and subgraph cropping; its follow-up JOYO \cite{You:2021wl} proposes to learn the augmentation priors via bi-level optimization.
GCA \cite{Zhu:2021gx} proposes adaptive augmentation that better preserves important semantics and structures of the underlying graph.
SimGRACE \cite{Xia:2022gd} eschews the need of explicit augmentation;
\citet{Trivedi:2022de} propose content-aware augmentation to avoid corrupting task-relevant information.

\end{document}